# Envisioning a Human-AI collaborative system to transform policies into decision models

**Vanessa Lopez** [a], **Gabriele Picco** [a], **Inge Vejsbjerg** [a], **Thanh Lam Hoang** [a], **Yufang Hou** [a], **Marco Luca Sbodio** [a], **John Segrave-Daly** [b], **Denisa Moga** [b], **Sean Swords** [b], **Miao Wei** [b], **Eoin Carroll** [b]

[a] IBM Research Europe, Dublin, Ireland
[b] IBM Watson Health, Dublin, Ireland

**Abstract**

Regulations govern many aspects of citizens' daily lives. Governments and businesses routinely automate these in the form of coded rules (e.g., to check a citizen's eligibility for specific benefits). However, the path to automation is long and challenging. To address this, recent global initiatives for digital government, proposing to simultaneously express policy in natural language for human consumption as well as computationally amenable rules or code, are gathering broad public-sector interest. We introduce the problem of semi-automatically building decision models from eligibility policies for social services, and present an initial emerging approach to shorten the route from policy documents to executable, interpretable and standardized decision models using AI, NLP and Knowledge Graphs. Despite the many open domain challenges, in this position paper we explore the enormous potential of AI to assist government agencies and policy experts in scaling the production of both human-readable and machine executable policy rules, while improving transparency, interpretability, traceability and accountability of the decision making.

## 1 Introduction

Governments regulate a wide range of sectors, with extensive policies, often in the form of coded rules. For instance, policies describing eligibility criteria for healthcare or social benefits, or port regulations for ships (based on load, size, dangerous cargoes, etc.). Automating these business decisions is a time-consuming task that requires substantial input from domain experts, and drives significant spending, e.g., an EU Commission study (2019) found that the annual cost of complying with EU financial regulations is around EUR€11.3B.

Policy automation is essential to consistently deliver services at population-scale. For example, checking citizens' eligibility to social care services - *who is eligible for what, how much and when?*. The automated version (code) becomes the de facto, 'effective', policy, being what most citizens experience in everyday life. Therefore, ensuring that the coded rules faithfully represent the original policy intent is critical. However, policies are complex, and gaps, biases and errors may remain unnoticed in the journey from policy intent to business requirements, to coded rules, to integration with existing data. These increase the risk of failing to deliver necessary services to citizens in need of them.

To address this, a recent global movement known as 'Rules as Code' (RaC) (Mohun et al., 2020) envisages a multi-disciplinary approach, bringing together legislative drafters, policy analysts and software developers, to manually co-create policy and code (i.e., machine-consumable versions of rules). The simultaneous expression of policy in both natural language and code provides powerful feedback opportunities for humane policy improvement. Policy expressed as code can be validated, checked for faithfulness to intent, combined with population data to discover gaps or biases, and analysed for 'what-if' scenarios.

We see enormous potential for AI, NLP, Knowledge Graphs (KGs) and Decision / Business Rule Modelling standards to assist policy experts and developers realizing this challenging RaC vision. In this position paper, we explore how these elements can be combined to speed up development of executable policy decisions, while also improving interpretability and traceability of the resulting rules. We explore an initial approach to design a first-of-a-kind AI prototype, guided by a UI proposed by domain experts, in the context of eligibility rules for social services. We discuss some of the open technical challenges and propose directions to fulfil on this vision along the way.

## 2   Background and State-of-Art

Policy is difficult to understand. Before coders can implement rules to automate policy intent, business analysts need to interpret policy. This involves first understanding the bigger picture – *what the benefit is, who can claim for it and what it offers*, which can contain references to sections in other policies that the analyst needs to locate to get the correct interpretation. Second, a detailed analysis is carried out to document business requirements and group 'rules', which makes them easy to follow. These 'rules' are not yet executable, they are typically documented in spreadsheets, and expressed in pseudo-code or simpler Natural Language terms. Third, business analysts need to identify the data needed for a rule to fail or pass. That means working out the conceptual model of the data, i.e., what entities and attributes need to be captured to substantiate eligibility decisions. The rules and data expressed in pseudo-code are then coded by the developers, according to the particular rules engine and legacy systems. Examples of rule engines are OpenFisca.org, IBM SPM (CER, nd), DROOLS (Drools, nd), and OPA (Oracle Policy Automation, nd).

Unsurprisingly, this long, multi-step translation from policy analysis to coded rules generates the following challenges highlighted by a recent OECD report on RaC (Mohun et al., 2020):

- Accountability of decision making - there is a need for better collaboration tools and the use of standards to document shared understanding, enabling business analysts to validate the implemented rules are true to the policy intent.

- Transparency and interpretability - needed to be able to explain, see and influence which rules are being applied, and on what policy basis.

- Traceability and maintenance - it takes too long to implement policy updates due to the lack of traceability from coded rules to the originated policy text and intermediate artifacts. During this time rules may be out of sync with policy, resulting in incorrect eligibility decisions.

Current approaches to achieving RaC range from manual coding by multi-disciplinary teams, examples include New Zealand Better Rules Discovery (2018), OpenFisca.org, Canada's Discovery project (Code for Canada, 2020); to approaches using NLP to assist policy experts in code-generation from a (controlled) natural-language representation of legal knowledge - explored by DataLex AustLII's (Greenleaf et al., 2020). Others are adopting as momentum grows.

Recent open-source standards, such DMN (Decision Model and Notation) published by (OMG, 2021) can be used to represent decision logic and dependencies, decoupled from business processes. DMN is designed to be understandable by business and technical users alike. Business analysts can model the rules that lead to a decision in easy-to-read tables that can be executed directly by a decision engine (Redhat DMN, nd), making DMN models fully executable. DMN builds on the notion of boxed expressions to represent the decision logic, and Decision Requirements Graph (DRG) to represent the dependencies between decisions, input data and other supported forms of business knowledge sources and reusable functions. The most widely used boxed expression is a Decision Table, which contains the logic to define how an output is determined from the inputs. The decision logic can be expressed in the standardised rule-based declarative language FEEL (Friendly Enough Expression Language).

SBVR (Semantic Business Vocabulary and Business Rules) is another relevant standard by (OMG, 2015). SBVR consists of a vocabulary of noun and verbs, and rules defined using terms in this vocabulary, based on a set of modal operators (e.g., necessary, possible, obligatory, permitted) and quantifiers. Similarly, the ODRL Information Model defines a semantic model for permissions and obligations statements formally specified using UML notation (W3C ODRL, 2016)

Generally, SBVR and ODRL differ from DMN as they do not consider inputs and outputs or document level dependencies between decisions. W3C SHACL (Shapes Constraint Language, 2017) can be used to model concrete regulatory requirements as constraints. Catala (Merigoux et al., 2021) is a programming language for law that also aims to support lawyers and programmers through a common executable specification.

Although business decision modelling has seen a surge of interest, limited research has been conducted regarding using AI to support their extraction from text. Approaches for rule acquisition by generating SBVR from text have focused on sentence-level rather than document-level rule generation, e.g., SBVRNLP (Bajwa et al., 2011) creates sentence-level normative SBVR rules using a rule-based algorithm for semantic analysis with English-language specification of business rules and UML model as input; BuRRiTo (Chittimalli et al., 2019) classifies single sentences as either a "rule sentence" or noise with a trained trigram language model and analysing the dependency tree. Other related work discusses the transformation of text to DMN. Text2Dec (Etikala et al., 2021) uses text mining to semi-automatically build a DRG, extracting decisions dependencies at a paragraph level, however they consider only edges, and rely on the assumption that the description is sequential, contains only one main decision and no irrelevant information. Another approach is discussed in DMNNLP (Garcia et al., 2021), where DMN decision rules and tables are extracted from single sentences using pattern-based text mining.

## 3 Vision and research challenges

We propose to leverage state of the art AI/NLP methods, KGs and standards to help analysts create decision models *directly* from complex policy documents with less effort. This is an introductory position paper on the challenges of building decision making systems on policy documents. To this end, we explore, through an illustrative example, how different AI techniques can be used in a collaborative fashion to identify eligibility rules and input data in policy text, and suggest decision model elements to represent them.

To support *fit-gap analysis*, KGs can be used to ground decision model elements to existing data when possible, promoting reusability when modeling. This speeds up analysts' work to identify similar pre-existing rules, codes and conceptual data needed as input to check eligibility (from this or other related policies). Keeping a tight link between the decision models, the originating policy, and conceptual (schema) data, facilitates models that are amenable to code generation and ensures automatic traceability.

Modelling standards are key to provide common understanding/representations and realize the vision of an AI-Human collaborative system. They facilitate decision rules and data types to be modelled in a familiar way, empowering the collaboration between business analysts and developers in maintaining them faithful to policy intent over time. Automatic translation of policies into decision outputs or code is not a goal - we see it as neither feasible nor ethical in complex, sensitive domains like health and social benefits. In this proposed prototype, we use DMN as it is widely adopted and well suited to code generation, and existing UIs can be reused for users to manually create fully executable DMN models (Redhat DMN, nd). Nonetheless, the output for the AI tech could be adapted to different or complementary emerging standards that may be better suited to other policy domains.

Applying this to our chosen policy domain (social benefits) we identify key following challenges to be addressed:

- The need for a corpus of labelled data to train NLP models in the chosen domain. This varies greatly across policies (e.g., child welfare, income benefits, food stamps).
- The need for cross-sentence, cross-document analysis to identify decisions, conditions, acronyms, etc., within and between policies.
- The need to identify relevant, pre-existing data and rules, for fit-gap analysis (and to avoid creating duplicate/divergent data and rules e.g., for checking citizenship/residency)

## 4 Methods and research directions

### 4.1 Identifying steps and AI techniques that assist user workflow

Together with a group of experienced social program business analysts and rules developers, we co-designed a User Interface (Figure 1), to support users accelerate the production of executable DMN models from policy documents.

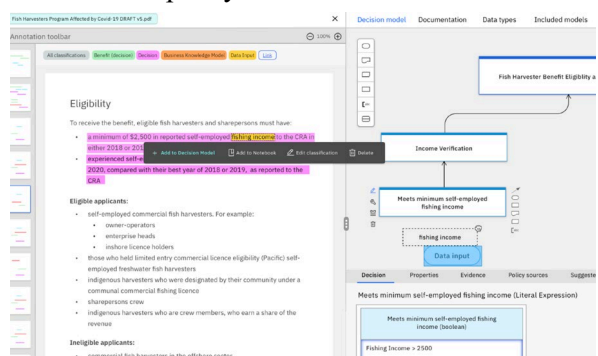

Figure 1: Envisioning an AI-assisted UI for rapidly building DMN models that are grounded in policy and existing data. AI services provide suggestions to users to model relevant text fragments in DMN, such as Decision Tables, or highlights dependencies to input data or outputs from other decisions.

The UI ties three key components together: 1. The set of user selected policy documents (left panel); 2. the eligibility decision under construction, as a DMN model (center panel); and 3. any existing data or rules that appear relevant to (or similar to) the currently-selected element in the DMN model.

The user uploads relevant policy documents. These are automatically pre-annotated to highlight (a) fragments containing eligibility criteria and (b) the terms within them that likely correspond to data elements – e.g., 'income' and attributes 'in 2018'. This requires NLP models that can identify fragments describing eligibility criteria, as well as NER models to identify relevant entities and relations. These are described in **Section 4.3**.

The user drags/drops highlights of interest in the policy onto the DMN diagram canvas, where they will model the eligibility decision using DMN elements. If an entity highlighted is dropped on the canvas, the user is offered relevant suggestions from similar existing datatypes. This assists the analyst in understanding which datatypes and codes are already in production use, and whether they 'fit' this use case. This requires a domain KG to flesh out policy and data requirements (entities and relations) by integrating knowledge from existing systems. This is described in **Section 4.2.**

If a sentence dragged from the policy relates to eligibility criteria, the user is offered relevant suggestions, such as creating a Decision Table. When there isn't enough labelled data linking text to DMN elements to train a classifier, these suggestions need to be driven by heuristics that map text structure to appropriate DMN elements. This requires deep parsing to analyse sentences and partially fill elements (e.g., a Decision Table with input–antecedent- and output-consequent- data types and values, or/and an S-FEEL expression). This is described in **Section 4.4**.

Closely related to this are suggestions that the system can offer relating to the DMN structure (DRG) and decision labels. These require models that analyze the discourse across conditions within and between sentences, such as income conditions or required participant information, and language models to detect policy fragments similar to a selected DMN element and that may contain complementary information, even if in different policy sections. This is described in **Section 4.5**.

In all of this, the user is in full control – able to ignore suggestions, add their own DMN elements, etc. Unused suggestions and user-created DMN additions are all potentially usable forms of feedback for improving the underlying models.

An example from Canada Fish harvesting benefit policy (2021) and a simplified DMN for the eligibility criteria expressed in the text is presented in Figure 2 for illustrative purposes.

### 4.2 Domain Knowledge Graph building

Knowledge Graphs can be used to capture and normalise domain information from specific rule engines, while abstracting from the particular engine implementations and data infrastructure.

To support the annotation of relevant elements in policy text, we propose collecting knowledge, representing entity types relevant for eligibility, such as existing rules, datatypes, attributes and/or codes, and the relations among them, in a KG. As an example, our current KG contains 1,040,271 statements and 164,130 entities, which have been automatically extracted from an eligibility rule engine product (CER, nd) implementing 514 federal eligibility rules, 257 datatypes/ attributes, 2,203 code tables and 22,261 code values. Even if this KG may be ambiguous and will not include all elements needed to model a new policy, such the fish harvesting, we propose to use it to support the NER by matching policy text to domain datatypes (e.g., residency status) and codes (e.g., nationality).

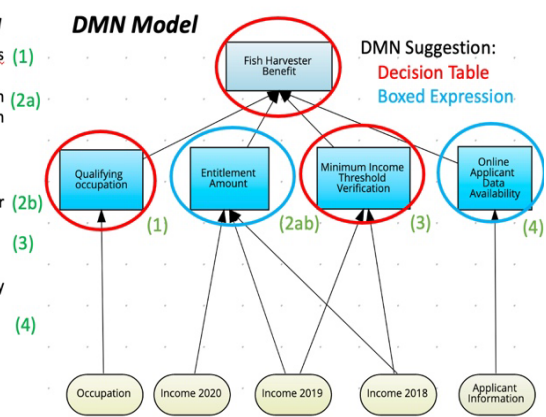

Figure 2: An extract from (Canada Fish harvesting benefit policy, 2021) and a simplified user created DMN. The DRG represents dependencies (edges) between the input data (represented by rounded ovals) and 4 boxed expressions, corresponding to decisions (represented by rectangles) described in the text: (1) occupation criteria for the beneficiary; (2ab) entitlement amount over income decline; (3) minimum income requirement; (4) required applicant information.

For example, if an entity highlight from the policy is dropped on the canvas, the user is offered relevant suggestions from similar existing types map in the KG – e.g., dragging a term like "fishing income" might suggest related KG types to represent 'income', including datatype attributes such as the *period start / end date* and the *income type* - e.g., 'fishing', if already defined in existing code tables (f*it-gap*).

An open challenge when using a domain KG extracted from existent rule engines to support the fit-gap analysis is the heterogeneity across policies and the ambiguity in the KG. This is emphasized by the presence of noisy, redundant, and duplicated legacy rules and data. Moreover, often the rules in existent rules engines are not linked to the specific policy text from where they were originally coded.

### 4.3 Zero-shot entity and relation extraction

To overcome the lack of labelled data and tackle the cold-start NER, we suggest the use of a zero/few-shot approach that leverages the domain KG to identify mentions of entities and relations in text that may be relevant as input data or part of eligibility rule conditions.

Figure 3 shows an example of links between mentions in the text and KG entities, using a simple NER and linking approach available in (zshot, nd), based on large pretrained language model (e.g., BERT), with an architecture similar to BLINK (Wu, 2019). In a first step, mentions are identified using frameworks such as Flair (Akbik, 2018) and Spacy (Honnibal, 2017).

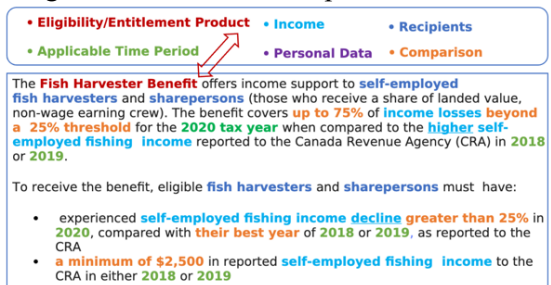

Figure 3: Domain entity annotations and types for a subset of the fish harvesting policy

Each mention is then encoded in a dense space, together with its context (left and right part of the sentence, excluding the mention). Independently, each entity in the extracted KG is encoded in the same dense space together with its context (e.g., entity description if available or nearest nodes links/description in the KG). This bi-encoder architecture can be used to link mentions in the text to entities in the dense space, using an approximate nearest neighbor search. Links with highest score indicate an entity match with good confidence.

To improve the NER further and possibly predict new (unknown) entities relevant for constructing data and rules, we need to leverage the DMN created by users in a feedback loop with supervised learning techniques.

### 4.4 DMN suggestions

Given previously annotated sentence and deep parsers, such AMR (Hoang, 2021), heuristics can be applied to suggest how to formalize it and the appropriate DMN elements to do so, based on the entities involved and the conditions. For example, if the text points towards a discretization of the possible values of inputs and outputs of a decision, it suggests modelling it with a Decision Table.

Figure 4 illustrates how a sub-condition on the minimum income threshold can be turn into a DMN Decision Table (lower part). AMR parsing (upper part) detects the main linguistic characteristics in sentences, such as co-reference, modality, quantities negation, coordinated or subordinated clauses. AMR reduces linguistic variations with respect to dependency trees, which makes it easier to extract antecedent and consequent, the decision logic between conditions (and/or), the outcome type, and data inputs for a decision. Using a few heuristics on top of the sentence level deep parsing, one could represent the semantics of the second sub-condition "[minimum income of $2,500]" as "[income > $2,500 in 2018] OR [income > $2,500 in 2019]".

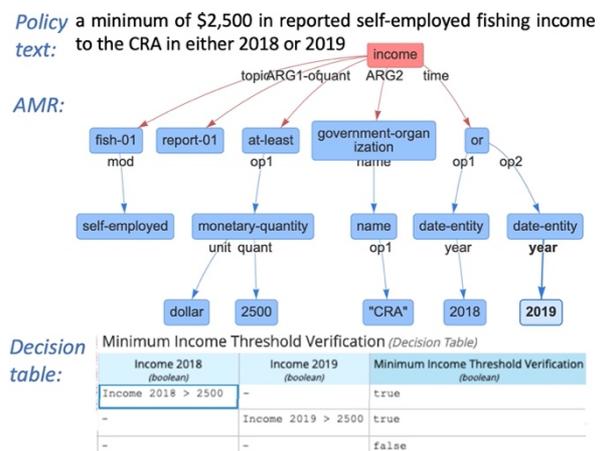

Figure 4: Example of AMR (Abstract Meaning Representation) for a sentence and its corresponding Decision Table consisting of columns representing inputs & outputs of a decision. Each row represents a rule with of 1 or more conditions and a conclusion.

If, for example, the text expresses more complex decision logic that is a chain of decision logic expressions (e.g., using a small subset of standard FEEL grammar operators such as arithmetic operators, unary operators, interval), decision tables or function invocations, then heuristics can suggest context box. For example, Figures 5-6 shows a more complex context expression, it leverages several literal expressions (context entries) that are used to derive the end result. These are difficult to extract from text in that form and therefore manual input from the business analysts is needed (human in the loop).

Heuristics are needed in the cold-start scenario to offer suggestions that support the user process to manually build the decision elements, linked to the policy text they are extracted from. Over time, these can be used as a valuable corpus of labeled data to train ML prediction models on policy text.

Figure 6 This DMN context expression box is used to model the entitlement amount calculation that corresponds to two related text inputs "The benefit covers up to 75% of income losses beyond a 25% threshold for the 2020 tax year when compared to the higher self-employed fishing income reported to the Canada Revenue Agency (CRA) in 2018 or 2019" and "To receive the benefit {..} must have: experienced self-employed fishing income decline greater than 25% in 2020, compared with their best year of 2018 or 2019". It leverages several literal expressions as values for its four context entries.

Figure 5. This DMN context expression box is used to model the online applicant data availability. This context expression box leverages three literal expressions (as values for its three context entries) to group Applicant (participant) Related Information from a conceptual perspective. The end result (*boolean*) verifies that all literal expression conditions are satisfied. It corresponds to the policy text "Individuals applying for the Fish Harvester Benefit using the online application need to ensure that they have the following information available: Applicant's Social Insurance Number (SIN), {..}"

### 4.5 Discourse model for decision elements

Different business analysts might come up with different structure to present conditions in their concept models, but one of the major challenges is to extract the discourse between decision elements such "*who is eligible for what benefit, under what conditions and for how long*" from the policy text, which looks similar to template-based information extraction (Nathanael and Jurafsky, 2011).

The difficulty lies in how to extract and present conditions and their logic structures. For instance, in the example in Figure 7, "[income decline greater than 25%] AND [minimum income of $2,500]" (elements marked in orange) is a valid condition expression, which contains two sub-conditions and their logic relation is "AND". These two sub-conditions are expressed as parallel elements (clauses/phrase) in the policy text. Although prescribed guidelines for formulating conditions do exist for families of legislation, they may not hold in every document, and the same semantics can be expressed in different linguistic structures, such as:

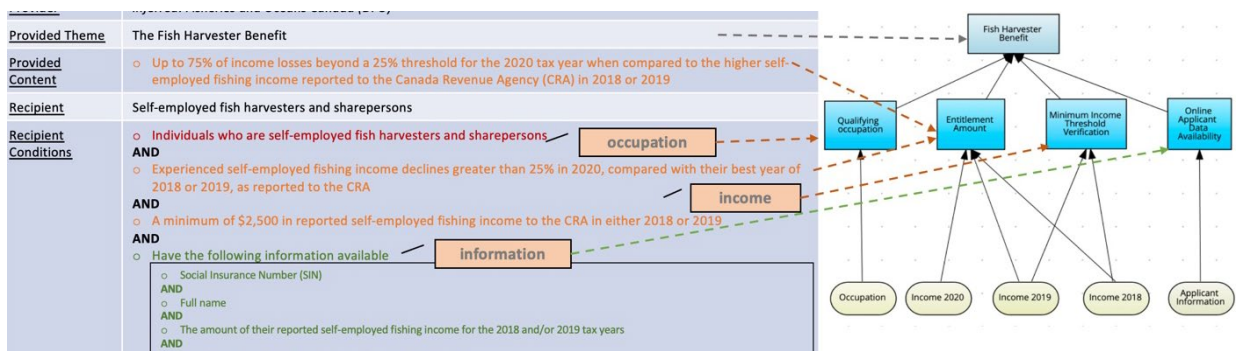

Figure 7: The table on the left shows discourse annotations and decision-based labeling based on a template expressing the semantics of decision rules in the social care eligibility domain. It needs to recognize the logic between the text fragments, e.g., OR/AND relations, and it can cluster extracted conditions into different groups, suggesting a short name (topic) for each group. On the right the DMN Decision Requirements Graph (DRG).

```
"To receive the benefit, eligible fish
harvesters and sharepersons, who have a
minimum of $2,500 in reported self-employed
fishing income in either 2018 or 2019, must
have experienced fishing income declines
greater than 25% in 2020, compared with their
best year of 2018 or 2019"
```

Furthermore, each sub-condition in the above example contains more atomic or fine-grained conditions. To support downstream applications, such as suggesting DMN elements discussed in Section 4.4, one should combine these atomic conditions in a meaningful way to fully express the semantics of each sub-condition.

In general, it is not straightforward to model these atomic or sub-conditions and their logic relations using the current computational linguistic models, such as Rhetorical Structure Theory (Taboada and Mann, 2006). Moving forward, we need to develop representation models and build corresponding corpora to support developing AI models for eligibility policies.

In addition, it needs to cluster conditions into different meaningful groups at document level. For instance, in Figure 7, one can cluster the conditions into three groups: "occupation", "income", and "information". For instance, the labeled condition groups in Figure 5, representing relevant decisions found within the text, are well aligned with the decision boxes in the DMN structure diagram.

A useful task is to detect text fragments describing similar decision information and calculations that can be reused in more than one place. Taken together, these fragments often provide a complete description of a rule that may not be fully described in any individual section alone. Transferred embeddings from pretrained language models, such as Sentence-BERT (Reimers, 2019) can be used to some extend to evaluate the similarity between two fragments, considering them similar if the cosine similarity over the two embedding vectors is greater than a predefined threshold. For example, the policy fragments (2a) & (2b) in Figure 2 define the entitlement amount and eligibility criteria based on the same calculation on income losses / declines.

## 5 Summary and future directions

We propose an initial guided AI-assisted approach to envision how foundational language models, KGs and standards could be combined together to streamline the collaboration among experts (policy analysts, developers and even open-source communities), from analyzing heterogeneous policy documents to derive computable decision models more effectively. The use of standards and KGs shift the focus towards policy interpretability and validation of its digital expression, which are even more important as the use of AI/NLP models increases. In turn, we expect human validated decision models linked to text will bootstrap the production of better AI models for this domain.

Our aim is to outline the challenges and research directions, where the NLP community needs to advance the state of the art to fully support this vision. AI techniques could be leveraged in different ways that the ones proposed here, however, the AI should be as non-intrusive as possible, and further research is needed to capture feedback from users as they build decision models – to use as ground truth for advancing legal document understanding and discourse, and develop better techniques for cross-section and cross-document capabilities (e.g., to identify rules describing the same benefit across-paragraph and documents).

At the same time, we are working with stakeholders to ensure adoption and deployment of future versions of our prototype into their workflow. Using standards to support users to translate legal text into decision models that are executable, even if a mostly manually process at first, might facilitate the creation of a consistent corpus of labelled data, linked to the policy text, that we could leverage to promote further research on AI-assisted policy automation.

Advancements are needed in improving document-level NLP techniques, building novel evaluation benchmarks for entity / relation extraction and identify dependencies among decisions and eligibility conditions from document-level policy discourse, as well as, designing user-based evaluations with industry practitioners engaged in delivering a RaC framework, in order to make the development of this kind of emerging collaborative systems a reality, at least for those policy decisions that can adequately be represented and automated with modeling standards, such DMN or others.

## Ethics Statement

Open government and Rules as Code emerging initiatives to increasingly automate aspects of policy comes with great opportunities for fairness— enabling policy rules on eligibility to be applied consistently. When everyone consumes common rules, there is less room for misinterpretation and the Rules as Code paradigm enables more comprehensive and broader testing and monitoring. Thus, supporting governments and social care agencies to deliver services at population-scale.

The problem of encoding regulations in rules is harder than what we could cover here. Besides, not being currently feasible to fully automate the creation of formal representations, rules or code from policy without domain experts' substantial input and interpretation, we are not interested in doing so. Organizations that automate policy have a responsibility to ensure there are no "translation of intent" errors when translating from policy to code. Therefore, our vision is designing guided AI-assisted services, where a first-class design goal is the human ability to correct 'easy to interpret' rules that feel natural and understandable to policy analysts, who need to make sure that the executable models will have the desired effect intended by the policy written by policy makers.

The combination of rich semantics, NLP, AI and standards will have a significant role to support the Rules as Code vision - shortening the route (time and cost) from policy to interpretable and executable code, while improving transparency, interpretability, traceability and accountability among stakeholders (policy analysts and rule developers), which are needed anywhere that citizen coverage is at stake.

Decision models, such as DMN, can be understood and corrected by both technical developers and non-technical, policy experts and analysts, empowering collaboration among business analysts, rules developers and/or even the open-source community through standardization and common understanding representations. There is a surge of interest on exploring the use of standards in the eligibility domain, we believe they also bring significant potential for leveraging AI/NLP capabilities and human expertise to provide suggestions to improve the efficiency to create formal representations of policy decisions, support detecting errors and gaps between policy and the documented decisions, maintaining and updating the models when policy changes are introduced, ensuring the code works as intended by the original policy(-ies) and enabling foundations for "what ifs" and fast feedback loops to understand impact of new legislation, omissions, or proposed changes to existing legislation.